# 14. The Use of Agricultural Robots in Orchard Management

Qin Zhang[1], Manoj Karkee[1], Amy Tabb[2]

[1]Center for Precision and Automated Agricultural Systems, Department of Biological Engineering, Washington State University

[2] United States Department of Agriculture, Agricultural Research Service, Appalachian Fruit Research Station

## 14.1 Introduction

The use of robotic or automated machines in orchard operations is associated primarily with insufficient labor availability and rapidly increasing labor costs in tree fruit production and is critical for improving yield of high-quality fruit with minimal dependence on seasonal human labor. Primarily, mechanized or robotic orchard management operations (after establishment of the trees) include pruning, thinning, spraying and harvesting.

Pruning is an operation to grow fruit trees into a desired shape for improving sunlight penetration into the canopies, supporting more effective (e.g. more human- and machine-friendly) orchard operations, and regulating location and amount of fruiting sites that can lead to improved yield and quality of fruit. Pruning is physically cutting and removing parts of a tree, such as cutting branches back and/or removing small limbs entirely following some guidelines. This operation requires workers have a specific level of knowledge and skill to perform the job well. Pruning is performed both in the winter (also called "dormant pruning") and in the summer. Currently, research and development of robotic pruning is focused primarily on dormant pruning, therefore, we will focus on dormant pruning for the remainder of this chapter.

Fruit trees' bloom often includes many more flowers than are required for producing desired yield and quality of a commercial crop. Crop thinning is a practice to manage/control crop load by removing a portion of the bloom and/or young fruit on the tree either selectively or non-selectively to improve the size and quality of the remaining fruit. Thinning is commonly performed during and shortly after the bloom period to reduce the number of set fruit using either chemical thinners or physical means. However, thinning can occur later in the season to remove poor quality fruit or those growing too close together to ensure good quality and size of the remaining ones using only physical means. Currently research and development of robotic thinning technologies is more focused on blossom thinning, and therefore, we will focus on this aspect of thinning in this chapter.

Pest control is an essential operation in orchard management to protect the yield and quality of fruit from various types of pests. One solution to achieve effective pest control is to spray recommended pesticides at the correct time and correct rate to the target area. Due to the confined environment in tree fruit orchards, spraying pesticides is inherently hazardous for the operators driving a sprayer. Robotic sprayers are probably the best solution to ensure operator safety and efficiency in pest control.

Fruit harvesting for fresh market consumption requires a large seasonal semi-skilled workforce which has created a critical labor-related risk of not having enough workers to perform this time-sensitive task. In addition, harvesting labor is already the most significant variable cost in tree fruit production. For example, labor represented 46% of variable production costs on a full production year of "Gala" apples (Gallardo et al., 2010). While vast efforts have been put on developing robotic solutions for fresh market fruit harvesting, the technology has achieved limited success, primarily due to inadequate *harvesting speed, efficiency,* and *harvesting induced fruit damage*. Research and development by both the public and private sectors continue to attempt to fill the gap of having a satisfactory solution for harvesting tree fruit crops using robotic machines.

Robotic management will only make operational and economic sense when all or most of the field operations are completed by means of corresponding robotic machines (or multi-purpose robotic machines) giving producers minimal dependence on seasonal human labor. As a complete harvesting process of tree fruit crops includes picking the fruit from the trees and loading it into bins, then transporting the filled bins out of orchards, robotic harvesting should include robotic transporting of the bins to complete the process.

The following sections of this chapter will provide an overview of robotic technologies for major tree fruit production tasks.

## 14.2 Robotic Pruning

As briefly discussed in Sec. 14.1, pruning involves cutting away of tree branches, and is performed for a diversity of horticultural and economic reasons, including: to balance the vegetative and reproductive growth of the plant, constrain the plant size, manipulate the canopy resulting in good fruit quality, size and yield, and set the optimal crop load for the next season (Schupp et al., 2017). Since fruit trees are perennial, pruning is a cumulative process, with the results of one year's pruning influencing the size and production of the plant for years to come.

Within fruit species, there are differences with regards to where reproductive flower buds, and consequently, fruit, will appear. As a result, pruning protocols differ by species. For instance, apple trees will flower and fruit on the last season's growth (called one-year-old wood). However, peach will only flower and fruit on growth that is two years old at the time of harvest. These differences mean that any autonomous system for pruning must take into account species-dependent differences so that the next season's crop is not inadvertently pruned off.

Trellis systems and tree training systems are also a major factor driving how pruning is accomplished in practice. We will not go into detail of individual systems here, but briefly mention that we will focus on systems that aim to automate selective pruning, where individual branches are precisely removed from the central trunk. Hedging, or mass removal via cutting bars, will not be covered, though it is sometimes an operation performed in orchards or vineyards.

Robotic pruning as interpreted in the literature thus far consists of perception of the shape of the tree, implementation of the pruning protocol, robotic navigation and cutting of the branch. Established computer vision techniques for estimating object shape with high accuracy typically rely on objects with easily identifiable features, such as corners or edges, a consistent or predictable image acquisition environment, or object smoothness. Dormant pruning occurs during the period when the trees are without leaves, the object is thin, and the data is acquired outside. These characteristics violate many of the assumptions of established techniques. For these reasons, new methods have been developed for perceiving tree canopies. Finally, the spacing of trees and permanent trellis structures restrict visual access to the tree canopy, and images can usually be acquired only from one side of the tree at a time.

While not autonomous, a motion tracker that works via the relative displacement of an electromagnetic field was used by Sinoquet et al. (1997) to model walnut trees and by Vougioukas et al. (2016) to model pear trees for automated harvest. The sensor can be used to record three-dimensional positions in space at branch junctions, which can be used to sparsely represent tree shape.

Time-of-flight (ToF) of light-based sensors have been most frequently used for sensing tree shape. Karkee and Adhikari (2015) described the use of a ToF camera to capture apple tree shape. The sensor has a small field of view, so a pan-and-tilt stage is used to acquire multiple images, which are combined in a data pre-processing step. Image processing and skeletonization steps are performed, resulting in the tree structure of trunk and primary branches. This system is also used in Karkee et al. (2014) to implement a pruning protocol based on branch spacing and branch length.

The Microsoft Kinect sensor is a commodity ToF sensor used in a range of vision tasks, and was employed by Elfiky et al. (2015) and Akbar et al. (2016b) for reconstructing apple tree shape as well. Approaches range from using the KinectFusion API (Izadi et al. 2011) that comes with the Microsoft Kinect 2 for surface reconstruction, and then skeletonization and feature extraction steps as in Elfiky et al. (2015), to an approach that estimates the skeleton and branch diameter features from the multiple depth images, which are then merged together.  In Akbar et al. (2016a), one Kinect depth image was used to estimate branch diameters. A dataset containing Kinect data from apple trees was described in Akbar et al. (2016b).

Medeiros et al. (2017) used a laser detection and ranging sensor, or lidar, for estimating tree shape for a pruning application.  In that work, a two-dimensional lidar unit was mounted on a rotary and linear stage to acquire large depth maps of the scene. Tree regions were modeled as cylinders, which was then used to extract various geometric features of the canopies including diameters of the primary branches.

A shape from inconsistent silhouette approach was used by Tabb (2013) and Tabb and Medeiros et al. (2017) to reconstruct the shape of fruit trees. Color cameras were used to acquire images of apple trees in an outdoor environment. Because automatic segmentation of scene between tree and non-tree regions is required, a background unit was used to block the appearance of other rows of trees.  While in Tabb (2013) multiple cameras were used in laboratory conditions, in Tabb and Medeiros (2017) a robot was used for data acquisition and is shown in Figure 14.1. Features computed include branch segment diameter, length, angle with respect to parent, and branch segment position.

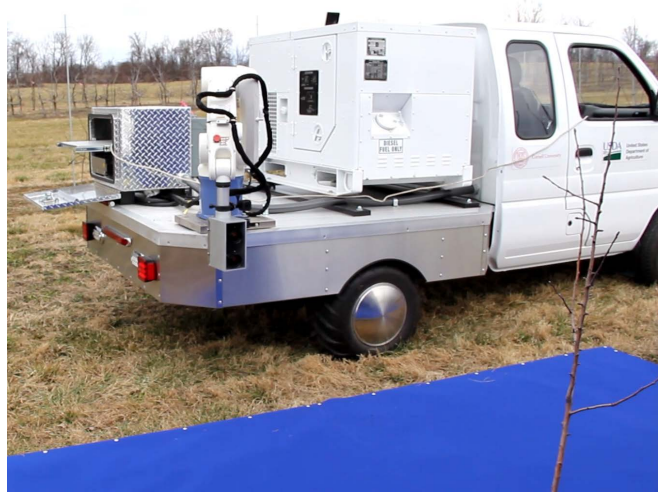

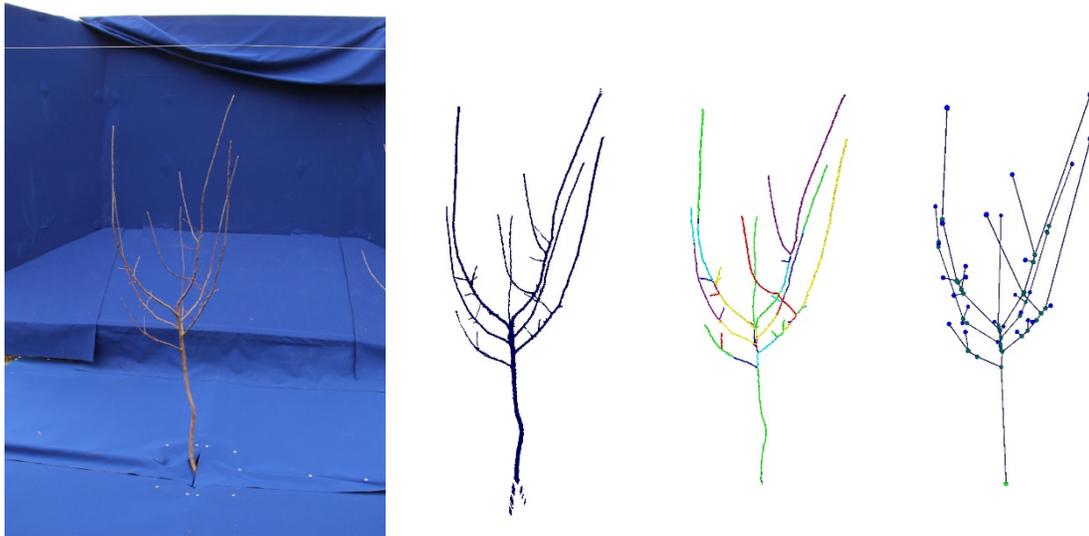

**Fig. 14.1. Top row: robotic vision system for estimating shape features of trees, Bottom row: image of tree, reconstruction generated by system, curve skeleton, and graph representation, all used to compute shape features. From Tabb and Medeiros (2017).**

Many research articles surveyed above estimated the branch diameter feature, some also computed the branch angle feature (Karkee et al., 2014; Karkee and Adhikari, 2015; Tabb and Medeiros, 2017). A subject treated in Schupp et al. (2017) concerns the development and testing of a quantifiable protocol for robotic pruning. The protocol, called the limb to trunk ratio (LTR), consisted of summing the cross-sectional area of all primary branches, and dividing this sum by the cross-sectional area of the trunk at a certain distance from the graft union. To achieve a target LTR, an autonomous system would select the branch with the largest cross-sectional area to remove, remove it, recalculate LTR, and repeat, until the target is achieved.

While the perception systems necessary for robotic pruning have been in development by various groups around the world, a major bottleneck to real-time implementation is computational time. For example, Medeiros et al. (2017) reported that data acquisition time was an hour for two trees. In Tabb and Medeiros' system (2017), the data acquisition time was 1.5 minutes per tree, and on average the run time was 8.5 minutes for small trees using a high-performance computer. None of these research activities considered navigation of a robot to cutting points. Further research and development is necessary for autonomous pruning to be a future reality.

### 14.3 Robotic Thinning

Fruit trees produce many more flowers than are required for a commercial crop with desired yield, fruit size and fruit quality. Thinning of flowers is performed to regulate the number of fruit that will set, and as a result, influences fruit size and quality. Thinning is also required to offset the tendency of some varieties of fruit crops towards alternate year bearing. The current practice is to thin the crop with chemicals, or manually. When chemical thinning, growers, extension specialists, and crop consultants use flowering counts to schedule thinning sprays as a component of crop load management. In apples, two models used in predicting thinning sprays require flowering as input (Yoder et al. 2012).

The small size of the target, the flowers, relative to the large size of the canopy poses some challenges for implementing robotics and automation solutions to this orchard task. Robotic and automated thinning has so far focused on flower thinning: including introducing elements of autonomy to mass removal systems such as string thinners with the development of robotic thinning arms and end-effectors. There also has been work on detecting flowers in color or multispectral images for two purposes: precise flower estimates as inputs to thinning models and three-dimensional estimation of canopy and flowers for robotic thinning.

String thinners were developed in the 2000s to mass remove flowers in high density fruit production systems (Baugher et al., 2010). String thinners are composed of a rotating spindle with plastic strings; the system is attached to a tractor which navigates through the rows in orchards. As the rotating spindle and strings pass by trees, flowers are removed. The angle of the string thinner, as well as the angular speed (revolutions per minute), can be adjusted to control the thinning efficacy. To develop sensing systems for automating some aspects of the operation, Aasted et al. (2011) used lidar to sense the canopy location for positioning the thinner appropriately in tree canopies trained in perpendicular V configurations. Gebbers et al. (2013) used a stereo camera to assess flower density on trees to support automated adjustments of thinner location and its spinning rate in real-time for optimal thinning in apple trees.

Computer vision approaches for detecting flowers in two-dimensional color images of fruit trees are concerned with either applications of chemical thinning, or as a module of robotic thinning. Some of these approaches will be discussed again in Sec. 14.4.1 in the context of robotic spraying. For such applications, images were acquired using hand-held or ground-based vehicles. One of the first efforts in this category, Aggelopoulou et al. (2011), aimed to adjust chemical thinning rates at the block level for a specific fruit yield in apples. Color images of trees while flowering were collected, as well as yield information, and correlation established between these two entities. The motivation in Hočevar et al. (2014) was orientated towards tree-specific management in apples; using tractor-mounted color cameras. They performed thresholding and morphological operations to estimate the number of flower clusters in acquired images. As part of a larger study, Underwood et al. (2016) detected almond flowers from color images using thresholding. Dias et al. (2018a, 2018b) detected apple and peach flowers from color images using convolutional neural networks and illustrated in Figure 14.2. Tabb and Medeiros (2018) and Dias et al. (2018c) are two public datasets with color images of fruit tree flowers, the latter with ground truths, for experimentation by the community.

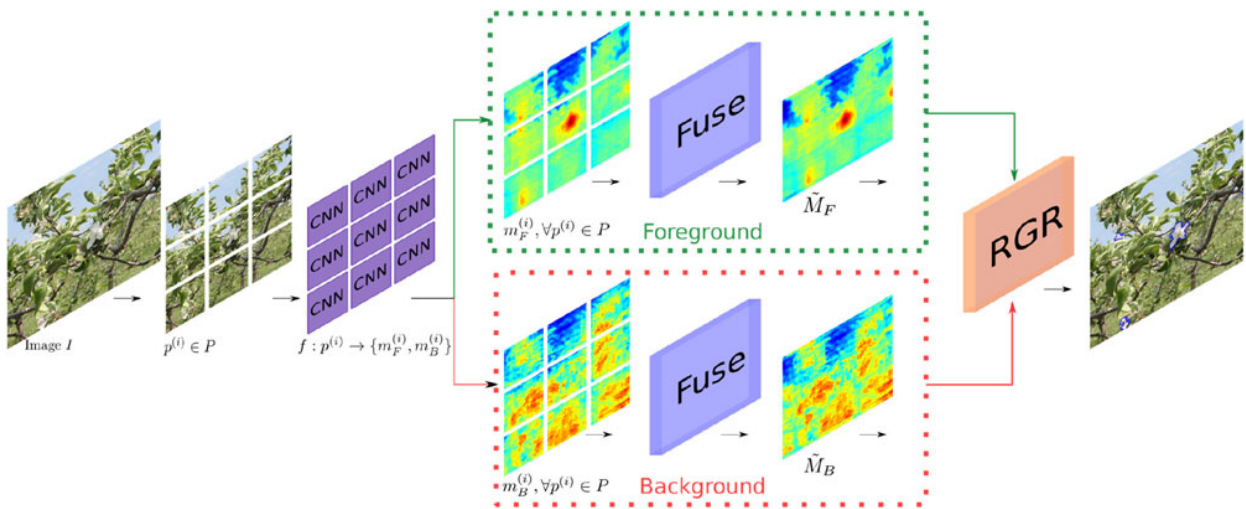

**Fig. 14.2. Schematic of a flower classification vision system from Dias et al. (2018b). Input images are divided and a CNN classifies pixels as flower class (foreground) or not (background). The region-growing-refinement (RGR) algorithm (see paper) is used as a postprocessing step.**

Some work made use of multispectral image sensors to detect flowers. For example, Horton et al. (2017) used multispectral information in the color and near infrared range captured using an unmanned aerial system (UAS) to detect flowers using contrast enhancement and thresholding methods in peach orchards. In peaches, flowers are visible before leaves, which makes an aerial survey approach feasible. Wouters et al. (2015) explored multispectral imaging as well for detecting many stages of pear flowers from a mobile, ground-based unit, with wavebands from the visible and near infrared range. Imaging occurred at night with artificial illumination, and canonical correlation analysis was used to create models to detect flower parts.

Robotic systems for flower removal will require knowledge of the canopy structure as well as the location of flowers. There have been efforts to investigate the three-dimensional reconstruction of canopies and flowers, as a component of a robotic thinning system, using the perpendicular V training system in peaches. Emery et al. (2010) developed a method to detect simplified peach canopies via structured light in a laboratory setting, and Nielsen et al. (2012) used a specialized stereo reconstruction technique, combined with flash illumination and nighttime operation, to reconstruct peach canopy locations and map flower locations in more realistic outdoor situations.

Other efforts in this area were concerned with the development of the robotic components for flower thinning. Yang (2012) in a master's thesis developed and evaluated a ¼ scale table-top robotic arm for the maneuverability necessary for thinning. Lyons et al. (2015) used a similar arm as Yang (2012) and demonstrated a brush end-effector for peach thinning that kept the branch damage at a minimal level.

Even when chemical thinning is performed, green fruit thinning is necessary and done manually in many fruit species after the threat of frost is past, and in some cases with a mechanical thinner (Miller et al. 2011). However, there are no efforts reported in the literature to date concerning robotic approaches to the green fruit thinning problem, neither green fruit detection nor its removal.

## 14.4 Robotic Spraying

As discussed in Sec. 14.1, spraying agro chemicals on orchard crops is one of the most important field operations to protect crops from various types of pests and diseases and to provide desired nutrients and other inputs such as plant growth regulators. Precise application of chemicals (i.e. right amount of spraying at right location and at right time, sometimes referred to as 3R) is essential to make sure inputs are applied effectively to achieve desired efficacy. Chemical application is an operation with a risk for adversely affecting human and environmental health, so health and safety regulations, such as limit on chemical residue levels on produce and the limit of chemicals movement away into neighboring fields, atmosphere, and/or water bodies, must be considered. Therefore, precision spraying has been and will continue to be a crucial technology in tree fruit production. With rapidly advancing computational power and sensing technologies, and revolutionizing artificial intelligence techniques such as convolutional neural networks (CNNs) or deep learning, robotic systems show promise for spraying agro chemicals to orchard crops with a higher level of precision than ever achieved.

Orchard spraying could use both ground and aerial robotics (e.g. targeted site-specific chemical application with UASs) in achieving desired level of precision. In addition to applying agro chemicals, precision spraying has also been investigated for artificial (also called mechanical) pollination of orchard crops. A robotic spraying system, in general, consists of a target sensing system, and a spraying control system to accurately and precisely deliver the right amount of materials to targeted areas in the canopy. In the following several sub-sections, recent advancement in various components and aspects of robotic spraying of orchard crops will be discussed.

For a targeted chemical application system, which attempts to match chemical application to canopy characteristics at different sections of the canopies without regard to individual canopy objects or parts, non-imaging sensors such as ultrasonic, infrared and LIDAR (Light detection and ranging) sensors have been used. These sensors can characterize canopies in terms of canopy depth, size, shape and foliage density at a relatively coarse resolution (depending on the number of sensors or scanning resolution used). For example, Jeon et al. (2011) used infrared sensors to detect canopy depth and other canopy parameters such as canopy volume. As illustrated in Fig. 14.3, the sprayer could therefore adjust the amount of chemicals applied to crop canopies in terms of the sensed canopy parameters to achieve a satisfactory result. Liu and Zhu (2016) tested a laser scanner for measuring shape and size of various types of tree canopies and achieved a coefficient of variance of up to 6.8% in measuring tree height and other dimensions.

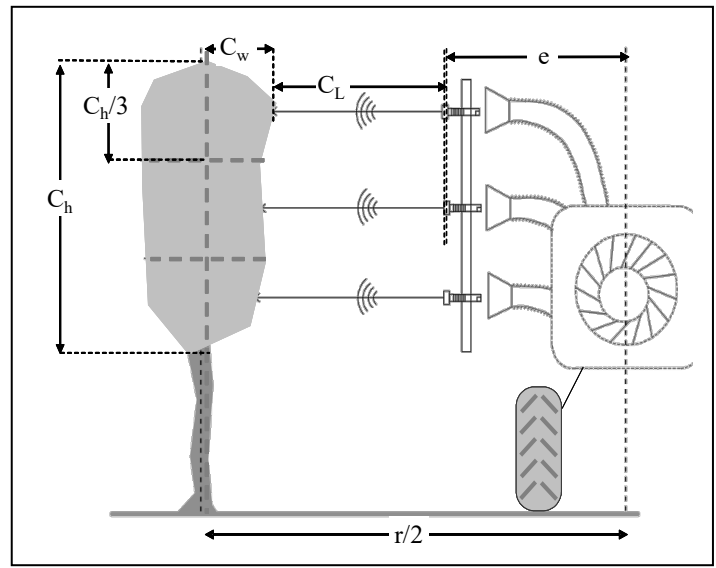

**Fig. 14.3. Ultrasonic sensing of various parameters for variable application of canopy inputs; r = inter-row spacing in orchards, $C_h$ = canopy height, $C_w$ = canopy depth, $C_L$ = distance to canopy surface from sensor, and e = distance to center of the row from sensor; adopted from (Jeon et al., 2011).**

If the precision needed is higher or inputs are desired to be targeted to specific canopy objects (e.g. flower, fruit, leaves, branches, trunks) rather than broadcasting to a specific canopy region, a machine vision system with the capability to detect and locate individual canopy objects would be necessary. For example, robotic pollination of cherry flowers requires accurate detection and localization of individual flower stigmas. Color cameras, multi- and hyper-spectral cameras and thermal cameras are some of the commonly used sensors for detecting canopy objects/parts whereas stereo vision cameras and laser scanners are the commonly used sensors for estimating 3D locations of objects. For example, Aggelopoulou et al. (2011) processed color images to estimate the number of flowers in apple trees and found it could be correlated with fruit yield with a correlation coefficient ($r$) of 0.78. A machine vision system was developed by Braun et al. (2010), constructing composite images from a sequence of color images and using a Bayesian classification to segment out crop foliage to determine spatial distribution of foliage in grape vines. In addition, contours and edges of flower parts (e.g. petals) have been used in detecting flowers in tree canopies (Hong and Choi, 2012). Multispectral imaging has also been investigated for detecting various parts of the canopies.

These machine vision systems utilizing geometric, color and spectral features of crop canopies suffer from the challenges of variable lighting, uncertainty of field environments, inherent biological variability of canopies, occlusion of objects, and variation in object shape, size, color and other parameters. To overcome some of these challenges in processing orchard images, deep learning techniques have been investigated widely in the last decade. Different variations of Convolutional Neural Networks such as Faster R-CNN **(**Ren et al., 2017**)** have been used leading to improved accuracy and robustness in detecting desired canopy objects and/or regions (Ahn et al., 2018).

As also discussed in Sec 14.3, 3D laser ranging and stereo vision systems have been used widely for localizing flower, fruit and other parts of the canopies in orchard environments (Emery, 2010; Aasted, 2011). Berenstein et al. (2010) used statistical measures, learning, and shape matching to distinguish and localize foliage and grape clusters in grape canopies and achieved an accuracy of 90% in distinguishing grape clusters from foliage. Dey et al. (2012) used structure-from-motion to create 3D structure of grapevine canopies, which were then used to detect and locate various parts of grapevines including leaves, branches and fruit. As mentioned in Sec 14.3, Nielsen et al., (2012) developed a sensor system to map the 3D structure of blooming peach trees, which was used to distinguish flowers from tree limbs using stereo-vision and color images. They achieved an accuracy of <1cm in locating flowers in the trees. Similarly, Gebbers et al. (2013) described a stereo vision system that used color contrast between blossoms, leaves and branches to detect and locate these canopy objects.

In many of these studies, two or more sensors were used for object detection and localization, which requires precise calibration and registration to project location information estimated by a 3D sensor onto objects detected using a 2D image. In recent years, commodity RGB-D (Red, Green, Blue-Depth) sensors, such as the Microsoft Kinect and Intel RealSense, have been increasingly popular in a variety of settings because of their low cost and the public availability of software development kits (SDKs). These sensors have the depth and color hardware components pre-registered to one another. Some groups have started using these types of sensors in orchard settings. Xiao et al. (2017) used a commodity RGB-D sensor to detect foliage using color information and to estimate leaf area density and average distance to canopy surfaces using depth information. Dong et al. (2018) used such a sensor and mapping algorithm to create three-dimensional maps of orchard blocks where depth information from two sides of the row is merged; traits autonomously measured include canopy volume, tree height, and trunk diameter.  Zhang et al. (2017) combined a commodity RGB-D sensor with CNNs to localize branches in apple trees.

Supported by an ultrasonic, infrared, laser, or LIDAR based canopy sensing system, it is possible to implement effective, targeted spraying by turning the nozzles on and off in terms of detected presence and absence of canopies in front of the nozzles. Gil et al., (2007), and Escolà et al. (2013) reported a few examples of using such automated technologies for spraying agro chemicals to the targeted zone at a desired rate on the canopy of various types of orchard crops. Zhang et al. (2018a) provided a comprehensive review on those technologies, and the number or the type of both the sensors and the nozzles installed on a sprayer can be optimized based on the crop types and canopy structures or architectures.

A precision spraying system developed by Xiao et al. (2017) could vary pesticide application rate based on detected 'leaf area density' and 'distance-to-canopy' using a RGB-D camera. This sprayer was tested in peach trees, apricot trees, and grapevines, which showed improved efficiency and reduced waste in spraying pesticides.

The ground spray systems discussed so far lack the capability to spray input materials to specific targets such at fruit or individual flowers. There have been some efforts at developing robotic solutions to spray targeted canopy objects or specific parts. One system targeted herbicide application to grapevine suckers, which grow at the base of grapevine trunks (Fig. 14.4, Kang et al., 2012). The system included a camera system to detect trunks and suckers of grape vines and a set of nozzles to spray chemicals to only targeted areas of the trunks to control sucker plants and insects, such as cutworm. The system was tested in wine grapes in Washington State showing a chemical saving up to 92% compared to a band broadcasting of chemicals with a regular sprayer. Berenstein et al. (2010) developed a machine vision-based automated system for precision spraying of hormones to fruit and pesticide to foliage without impacting non-target areas. A similar system was investigated by Oberti et al. (2016) for controlling powdery mildew in grape canopies. Tests in a greenhouse environment achieved an accuracy of 85% to 100% in detecting the diseased areas resulting in up to 35% saving in chemical use. Further research and development is warranted to make these systems robust and widely applicable to targeted flowers, fruit or canopy parts where the problem is detected. Research and development efforts need to focus on both sensing systems that can identify and locate the presence of specific pests or issues in crop canopies and a robotic system that can precisely apply desired amount of chemicals to target objects or canopy parts.

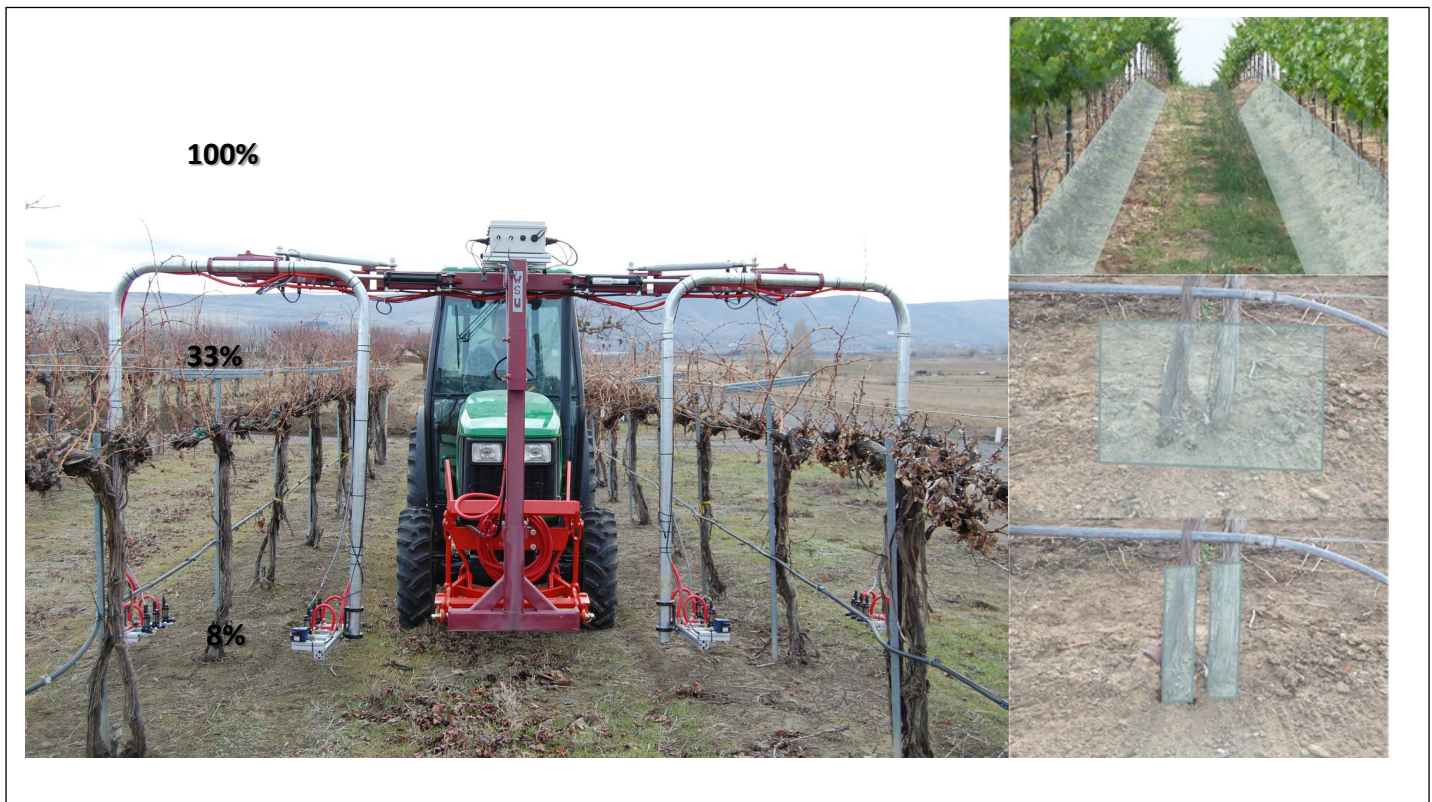

**Fig. 14.4: Left - A precision spraying system developed and tested by Washington State**

**University for targeted application of pesticides and other pest control materials to grapevines. Right – Comparison of amount of chemical used with a conventional broadcasting method and targeted, precision application to grapevine trunks.**

Other than the application of agro chemicals, artificial pollen application (also called mechanical pollination) is another important application of spraying technology in producing fruit, berry, nut and vegetable crops. There are primarily two approaches investigated for artificial pollination; aerial platforms and ground platforms. Aerial platforms include various designs of small UASs which often apply a bombing technology to spray pollens down onto the crop canopies from above. An innovative newer approach of aerial pollination is the use of a fleet of nano-aerial robots to mimic bee behavior as artificial pollinators. Berman et al. (2011) has developed and tested robotic bees that showed potential for pollinating crops when a swarm of those vehicles are used in orchards. Similar development of aerial robots has been reported by Abutalipov et al. (2016).

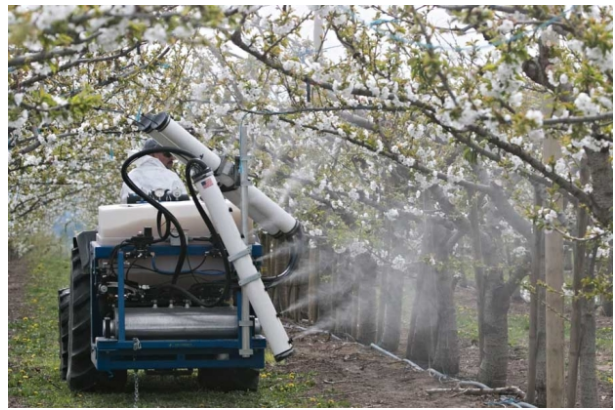

Artificial pollination using ground robots/vehicles is very like agro chemical applications using electrostatic sprayers. The machines are often designed to spray pollen suspended in a liquid or dust medium to target canopy areas. Different types of sprayers have been tested to optimize the type and size of nozzles, operating pressure, flow rate, carrier medium and distance to bloom so that the desired density of pollen could be sprayed to targeted canopy areas. For example, Whiting (2016) has tested an electrostatic sprayer for artificial pollination and found that supplemental application of pollen (suspended in a sucrose and boron solution) increased the pollen density on flower stigma by three-fold compared to natural pollination with bees (Fig. 14.5). Though there have been several systems evaluated for broadcasting pollen to target crop canopies including some commercialization efforts (e.g. a mechanical blower system commercialized by PollenPlus™, New Zealand), research and development of robotic systems (with manipulators and end-effectors) for selective pollination of a desired type and number of flowers have been limited. One example system was developed and tested by Barnett et al. (2017) in kiwifruit. The vision system included a camera and CNN-based image processing system for flower detection. The robotic machine included an autonomous platform and a spray manipulator with various degrees of freedom. The system was able to detect more than 70% of the flowers and over 80% of the flowers were pollinated with the robotic system.

**Fig. 14.5. An electrostatic sprayer retrofitted by Dr. Mathew Whiting and his team at Washington State University to apply pollens; being evaluated in a cherry orchard in Prosser, WA. Picture by (TJ Mullinax/Good Fruit Grower).**

## 14.5 Robotic Harvesting

Harvesting is a labor intensive and time sensitive operation in orchard crops. Scientists and engineers have been investigating robotic systems for harvesting various types of crops including apples (e.g. Rabatel et al. 1995), citrus (e.g. Harrel et al., 1989), kiwi fruit (e.g. Bazley, 2017), and cherries (e.g. Tanigaki et al., 2008). Unfortunately, so far, no commercial success has been achieved. In recent years, however, there have been heightened interest and

21-10

investment in research and development from both public and private organizations. With the latest development in sensing, computing and artificial intelligence technologies, robotic harvesting of fruit appears now to be a viable technology.

A robotic harvesting system for orchard crops consists generally of a vision system to detect and locate target fruit (and obstacles when needed), a manipulation system to reach the target fruit, an end-effector to detach fruit from branches and collect them onto a conveyance system, and a conveyance system to bring harvested fruit onto a container/bin (some of the components are depicted in Fig. 14.6). In this section, some of the latest efforts in various aspects of robotic fruit harvesting are discussed. In addition, an alternative method of automated fruit harvesting, a targeted shake-and-catch system, will be introduced. More detailed discussion on past efforts in fruit harvesting, conveyance and container filling technologies can also be found in review articles and book chapters published recently including Bac et al. (2014), Gongal et al. (2015), Zhang et al. (2016), and Karkee et al. (2017).

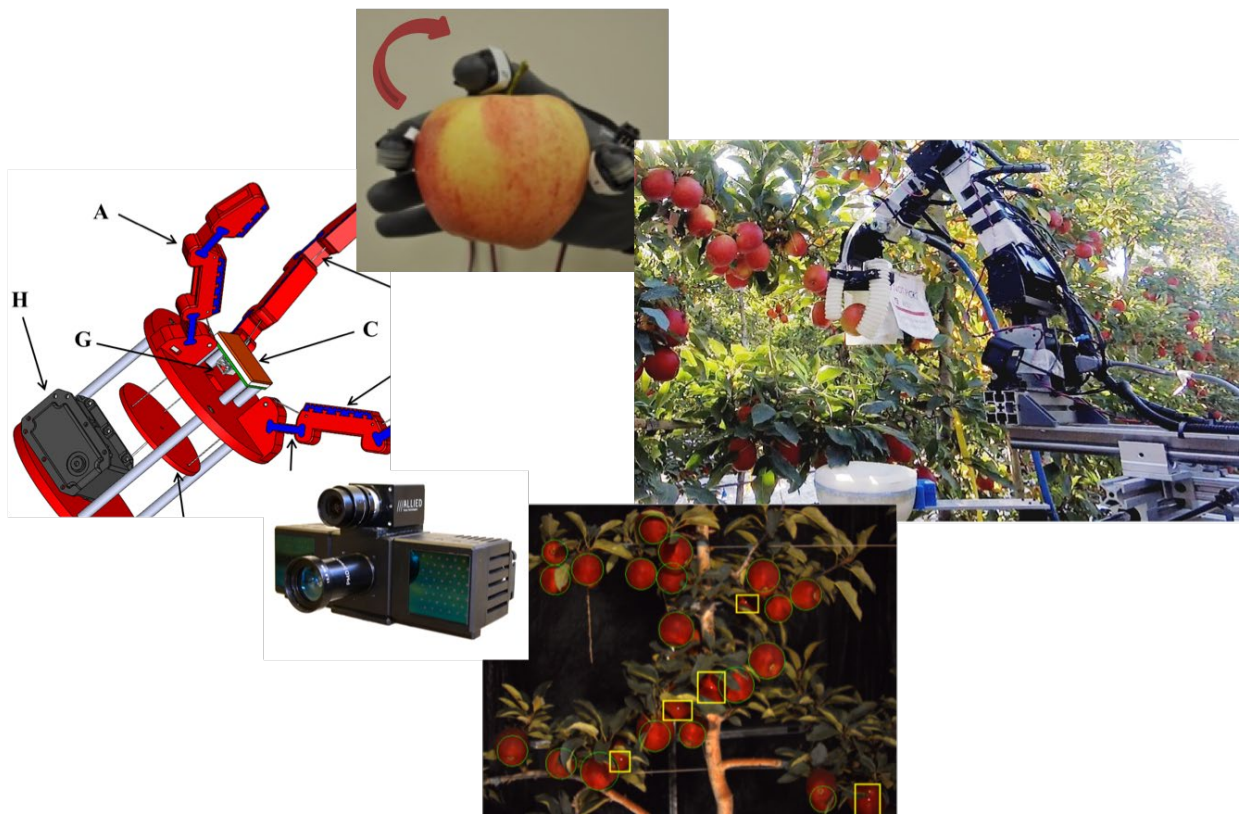

**Fig 14.6. A robotic harvesting system developed at Washington State University by a team led by Manoj Karkee and Qin Zhang (Silwal et al., 2017).**

The first step in robotic harvest is to detect fruit and estimate their 3D location in tree canopies so that an end-effector could reach target fruit and detach it from the tree. A wide range of studies can be found in detecting fruit and obstacles using specific features including, color, shape, size, edge and texture and using various thresholding and classification techniques such as Bayesian classifier and neural networks (e.g. Silwal et al., 2014, Tabb et al. 2006). However, these specific feature-based techniques have achieved only limited success due to issues including occlusion, fruit clustering, crop and canopy variability, and unstructured, uncertain and variable lighting conditions. To address the challenge of fruit clustering, Changhui et al. (2017) used a convex hull technique to identify individual citrus fruit and their centers (assuming circular fruit) in images with overlapped fruit clusters. Wang et al. (2017) proposed an image enhancement technique involving the wavelet transform and Retinex principle (McCann, 2014) to minimize the issues with variable lighting conditions.

In recent years, deep learning has been used successfully to address various challenges in fruit detection (e.g. Zhang et al., 2017; Chen et al., 2017). Stein et al. (2016) used deep learning networks for mango detection and localization whereas Halstead et al. (2018) used an end-to-end deep learning network to detect number and quality of sweet peppers. Chen et al. (2017) used a fully convolutional neural network (F-CNN) to detect and count oranges and apples with a mean intersection of union of 0.813 and 0.838 respectively for oranges and apples. Häni et al. (2018) also used CNNs to localize apple fruit from multiple tree varieties.

Fusion of color and three-dimensional information has also been investigated recently for multi-class canopy image classification. Tao et al. (2017) used support vector machines and a genetic algorithm to classify apple tree canopy images into fruit, branches and leaves. Similarly, Zhang et al. (2018b) used color and depth information collected by a Microsoft Kinect sensor to train a faster R-CNN, which showed better accuracy in classifying branches, trunks and background compared to the same without depth information.

Approaching a fruit is a critical step in robotic harvest, which primarily involves defining an optimal path to move the picking end-effector to the target fruit to complete the fruit detachment from the tree. Sequence or prioritization of fruit picking could be defined using the Travelling Salesman Problem (TSP), which can be solved using various algorithms such as nearest neighbor (Kizilates and Nuriyeva, 2013) that can minimize the total distance travelled, time taken, energy used or some weighted combination of parameters that define the performance of a robotic system (Silwal et al., 2017). Once the sequence is known, visual surveying could be used to approach the target fruit (Ringdahl et al., 2018). With visual surveying, approaching a fruit would involve detecting the fruit and its position (often in 2D image space) frequently using an end-effector-based imaging system and changing the manipulator joint positions such that the target fruit remains at the desired image coordinates all the time.

An alternative method to visual surveying is using a global camera system, by which a camera is mounted at a fixed position to take images at the beginning of a number of harvest cycles. Fruit position is then estimated at the beginning of the harvesting cycle for all the fruit available in the given field of view or the workspace. Once the initial or current position of the end-effector and final position (fruit location) are available, an inverse kinematics is used to estimate new positions for all the joints of the robotic manipulator that would move the end-effector to the desired final position and orientation. One challenge with this technique is to detect and locate fruit accurately at the beginning and to have an accurate calibration between camera coordinates and manipulator coordinates so that the end-effector could reach the fruit precisely.

Fruit detachment from the tree is the core in robotic harvest of tree fruit. It commonly uses a fruit picking end-effector to exert desired force and movement onto a fruit to detach it from the tree. There are different types of end-effector techniques being used in fruit picking. One technique is to detach fruit using mechanical end-effectors with a human-like hand and fingers. A soft palm is required to prevent the fruit from being bruised. Different designs have been used in a mechanical hand including different number of fingers and actuator to operate the fingers (Davidson et al., 2016). One of the ways fingers are actuated are electric motors running each joint in a hand individually, which requires a number of actuators per hand making it relatively slower, and more complex and expensive. Another way would be to use a tandem design (under actuated hand) where one central motor would drive all the fingers using a cable such that the hand can close to the target fruit confirming the fruit size and shape as the desired amount of force/tension is exerted in the cable (Davidson et al., 2016).

A mechanical hand may also use soft-robotic materials and pneumatic actuation (Shintake et al., 2018). Fingers in soft-robotic hands are designed with a hollow space inside the fingers with a wrinkled surface on outer side. As the hollow space is filled with air, the fingers close on the target fruit as the wrinkled surface expands outward due to pressurized air. This kind of fingers and the hand can operate much faster than the motor-operated, mechanical hands. In addition, soft hands provide a level of cushioning to the fruit being detached. However, generally, these fingers tend to be thicker than traditional counterparts, making it challenging to harvest closely spaced fruits or fruits in clusters.

Another type of end-effector being used is the scissors type which cuts the stem to detach fruit. For example, Bac et al. (2017) designed a four-fingered hand following the "Fin Ray principle" for grabbing the fruit and then mounted a pair of scissors above the hand to cut a stem. However, detecting and locating stems for cutting with scissors is a highly challenging machine vision task. To address this issue, devices such as a cup-shaped, conformal scissors that closes around a fruit to cut the stem irrespective of its position (Rosenberg, 1975) may be used. This technique is more suitable for fruit varieties with long stems. Once the fruit is detached and delivered using a robotic or mass harvesting system onto a conveyance system, they have to be conveyed, decelerated at the end of the conveyance, and collected onto a container. There have been various successful efforts in this area, details of which can be found in Zhang et al. (2018c) and Karkee et al. (2017).

**14.6 Robotic Fruit Transportation**

To implement robotic harvesting with a minimal seasonal workforce, it is desirable to use a robotic carrier to transport empty or full fruit containers/bins to and from the working zone. A functional robot carrier requires having basic functionalities of i) autonomous navigation; ii) intelligent in-orchard management; and iii) bin handling. Fruit bins used in the Pacific Northwest (PNW) region of the U.S.A. are typically designed to hold about 400 kg fruit during harvesting, transportation and storage. Therefore, two basic requirements of a robotic bin carrier are its capability of handling a load up to 500 kg and effective maneuverability in an orchard environment (Ye el al., 2017). Current robotic technology is capable of meeting this need by providing a robotic self-propelled fruit bin carrier system for transporting bins autonomously (Hamner et al., 2011).

The navigation functions of a robot bin carrier consist of; i) guiding the bin carrier with an empty bin into a tree alleyway where harvesting is performed; ii) moving to deliver and place the empty bin in the alleyway to an appropriate location in a harvesting zone; and iii) carrying a full bin out of the alleyway and delivering it to a repository station. A capable navigation system will require having both a GPS system for navigating the carrier between the repository and target alleyways; and a distance scanning system (either Lidar or ultrasonic sensors) for detecting the boundaries and obstacles within the alleyway where GPS signals are often compromised by trees (Ye el al. 2017).

An intelligent management system, including an automated robot control system, was used by Zhang et al. (2015) to autonomously plan and guide multiple robotic bin carriers in handling bins in the orchard to support efficient harvesting operations. This intelligent management system coordinates multiple robots working faultlessly and efficiently through making two decisions of i) which full bin should be picked up and returned to the repository; and ii) where it should carry an empty bin. To plan for efficient coordination in multi-robot management, Kalra et al. (2005) developed and validated an algorithm based on a market-based framework, which applied an auctions approach. This approach was found to be efficient for sequential management and implementation: once the intelligent management system makes a decision, it would proceed to complete its designated task and only makes another decision when it has finished its current task (Zhang et al., 2015).

The maneuverability of a robotic bin carrier concerns bin handling within confined fruit tree alleyways. Typical bin handling operations require the robot to i) load an empty bin at a bin repository station; ii) lift the empty bin and going over a detected filled bin on its way to place it at an appropriate location; 3) position itself properly to ready picking the filled bin on the way back; 4) pick the filled bin reliably and quickly; and 5) carry the picked bin securely out of the alleyway back to the repository (Ye et al., 2018). Implementing all the basic functions effectively and reliably in commercial orchard environments could be very challenging, mainly due to the confined working space in fruit tree alleyways, and the interference of randomly grown and deformable tree canopies.

Researchers found that use of a four-wheel-independent-steering system on wheeled field robots could effectively solve this problem (Bak and Jakobsen, 2004; Nagasaka et al., 2004). Such a steering system allows a wheeled robot to be maneuvered in four steering modes of two-wheel-steering (front or back Ackermann steering), four-wheel active-front-and-rear steering, spinning (the robot rotated about its geometric center), and crab steering (all the wheels have the same steering angles) as illustrated in Figure 14.7. This steering mechanism, when controlled properly, could achieve the maximum maneuverability of a wheeled robot in confined working spaces (Oksanen and Backman, 2013; Ye et al., 2018).

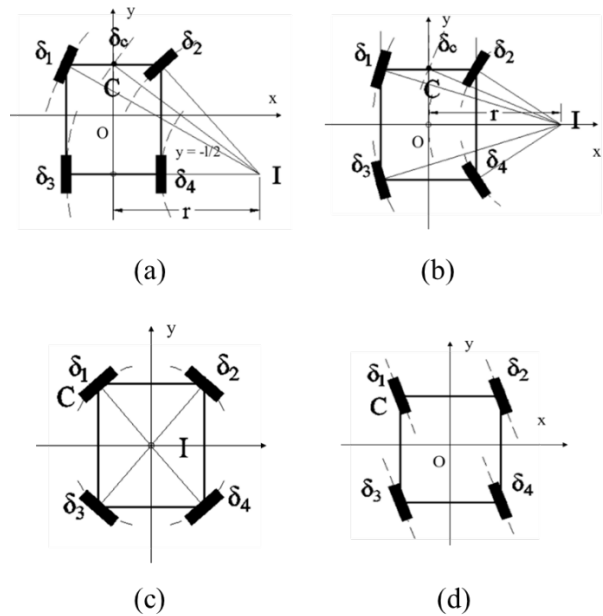

Figure 14.7. Four steering modes of four-wheel-independent-steering for wheeled robot: (a) Ackermann steering, (b) active-front-and-rear-steering, (c) spinning, and (d) crab steering

**14.7 Outlook and Summary Remarks**

After more than 50 years of persistent research and development of automated and robotic orchard machines, there seems to be an accelerated success in the past decade. Such an acceleration may be attributed to the substantial advancement in agricultural robotics enabled by powerful and robust computing, sensing and automation technologies created for other industries. Various companies around the world have shown interest in developing and commercializing robotic systems for crop production in recent years. Venture capital and other funding sources such as USDA-Small Business Innovative Research programs have also provided funding to encourage companies developing robotic solutions for production agriculture, including fruit and vegetable harvesting, weed control, and fruit tree pruning. A couple of examples include Agrobot S.L. (Huelva, Australia), which has successfully commercialized a strawberry harvesting robot (Agrobot.com), and Robotic Plus, a New Zealand-based high-tech company, which has introduced a robotic Kiwi fruit harvesting machine to the market (Bazley, 2017).

A major reason for the slow progress and adoption of robotic technologies to tree fruit production could be the remaining, specific technical challenges for agricultural robots. Among those challenges, one of the most difficult ones is the need for the robot to interact with deformable plants in varying operational conditions. Agricultural robots often need to handle targets that might move away from its detected position when the actuator starts to act on the target object. The plant deformation could be induced either by the natural stimuli or induced by robot-plant interaction, which could result in the end-effector missing the target or failing to form a required reaction to complete the designated tasks.

Another challenge to be solved is operational safety and efficiency. One basic requirement for any operation using a robot or other unmanned autonomous machinery is the safety of both the people working on sites with the robot and the robot itself. Production agriculture operations are generally performed in natural, varying and unstructured environments bringing additional challenges to achieve a sensitive, dynamic and robust job-site situation awareness. Robots' incomplete situation awareness may cause large safety hazards. Such incomplete situation awareness may also be attributable to current robotic systems' low efficiency to perform the designated tasks.

Economic concern always plays a big role in adopting new technologies for end-users in agriculture. Farmers often wish to keep the total cost of adopting a new technology at a level comparable to their current operations. For example, the average cost for growers to pay to seasonal apple pickers in the state of Washington, U.S.A. was up to $28 USD for a full bin (typical size of 1.2 × 1.2 × 0.6 $m^3$ with around 420 kg of fruit at full load) in 2017 (Bacon, 2017). While this target price for adopting robotic harvesting today could be very challenging, with the continuously increasing harvest labor cost (e.g. a rapid increase from $13 USD per bin in 2001 to $28 USD in 2017), the break-even time may come more quickly than anticipated. Nevertheless, keeping total grower costs, including the equipment purchases, maintenance, insurance, risks of crop loss caused by equipment reasons, and all the relevant operation costs are always important concerns for robotic operations in production agriculture.

To relieve farmers of the dependency on human labor in production, multiple robots are needed to perform different field tasks, during specific limited times, resulting in a very low equipment utilization efficiency. As for any other commercial product, agricultural robots will be manufactured and marketed in a standard initial setting suitable for as broad a market as possible to ensure the economics of manufacturing are sustainable when these robots are commercialized. The cost per unit time usage would be very high if they want to own and maintain the machines These economic scenarios set a difficult bar for many small farms to overcome in adopting robotic machines.

Another challenge is the adequate skill and knowledge required to implement robotic field operations effectively and efficiently. Cultural practices, field environments, crop varieties planted, and farming preferences or habits can vary widely for farmers from different parts of the world, or even between neighboring farms. This may lead to the need  for customized initialization to make operation-specific adjustments, and more importantly to create adequate robotic operation plans, before a robot product could be optimized for a particular end-user.

With a large percentage of aging farmers worldwide who lack knowledge of computing and robot technologies, a significant change in farmer and worker skillset will be needed.

Either technical or operational solutions could remove these challenges, however, more studies are needed to make such potential solutions practically usable. For example, if we could create a trustworthy situation awareness and adaptive mechanism with self-learning capability as a technical solution, it could be possible to remove the need for robotic machine initialization to meet user-specific operation conditions as well as make robotic machines able to react to detected situation changes similar to that of a human worker. Another possible technical approach is the adoption of a design philosophy of tractor-implement systems to build agricultural robots on a common platform providing power, sensing, task management, and control capabilities to support a fleet of interchangeable task-specific end-effectors for conducting different field operations such as harvesting, pruning, training, thinning and spraying.

While most farmers normally possess their own machinery to perform field operations, different adoption models could be worthy of investigation for robotized farming. For example, while possessing a fleet of robots could still be technically and economically feasible for some large farms, will a community co-op model, namely different farmers owning different robots and sharing use of those robots to reduce capital investment, be feasible? Or will commercialized robotic field operation services, sometimes termed contract farming, be another solution? Nevertheless, either option may create a new industry, namely agricultural robot support networks, and bring in professionals of agricultural robot support to rural areas.

Other than the progress in robotic technology and their application, the importance of developing some types of machine-friendly architectures of tree canopies can never be over-emphasized. Such machine-friendly architectures are obtained through training and pruning to convert randomly grown thick canopies to more uniform and accessible canopies. Over the last several decades, numerous different types of training and pruning systems have evolved around the world leading to many different types of more accessible tree architectures (Fideghelli et al., 2003). While today's robotic technology could find out ways to penetrate a randomly grown thick canopy of a fruit tree to perform some specific orchard operation, a more uniform and accessible canopy would simplify the mechanism of a robotic machine for performing those operations more effectively. It would not only reduce the price of the robotic machine at purchase, but more importantly could help to improve the effectiveness, efficiency, and robustness while potentially enhancing the safety of the machine being used in commercial orchards.

In summary, robotic technologies for tree fruit production have never been so close to being practically applicable. While there are challenges remaining to be solved before robots could be commercially adopted in orchard operations, it is expected that robots would make their way into some of the field operations in tree fruit production in the near future. Advancement in agricultural robotic technology may also bring in some important societal and economic impacts to rural, agricultural areas. For example, it might bring a demographic change in rural populations as new farming styles create a new industry to support robotic farming, which will bring in well-paid, year-around jobs both for supporting the farming and servicing of machines to replace close-to-minimally paid, seasonal field laborers. Such a change in the employment force will bring in more well-educated, young professionals to work in rural areas, which in turn will bring in more societal services and professionals, such as healthcare facilities, schools, retails and restaurant services leading to potential boom in the rural area economies.